\documentclass{article}

\usepackage{multirow}
\usepackage{geometry}
\usepackage{cite}
\usepackage{amsmath,amssymb,amsfonts,bbm,amsthm}
\usepackage{algorithm}
\usepackage{MnSymbol}
\usepackage{graphicx,tikz}
\usepackage{enumitem}
\usepackage{flushend}
\usepackage{textcomp,multicol}
\usepackage{xcolor,url,comment}

\usepackage{algpseudocode}

\let\oldReturn\Return
\renewcommand{\Return}{\State\oldReturn}

\usepackage[figuresright]{rotating}

\newtheorem{corollary}{Corollary}

\newtheorem{proposition}{Proposition}

\newtheorem{remark}{Remark}
\newtheorem{definition}{Definition}[section]

\newcommand\btab{\begin{tabular}}
\newcommand\etab{\end{tabular}}
\newcommand\bfig{\begin{figure}\centering}
\newcommand\efig{\end{figure}}
\newcommand\bfigs{\begin{figure*}\centering}
\newcommand\efigs{\end{figure*}}

\newcommand\Idle{\mathrm{Idle}}
\newcommand\bT{\mathbf{T}}
\newcommand\GT{\mathbf{GT}}

\usepackage{authblk}

\title{Improving Makespan in Dynamic Task Scheduling for Cloud Robotic Systems with Time Window Constraints}
\author[1]{Saeid Alirezazadeh}
\author[2]{Lu\'{i}s A. Alexandre}
\affil[1]{C4 - Cloud Computing Competence Centre (C4-UBI), Universidade da Beira Interior, Rua Marqu\^{e}s d'\'{A}vila e Bolama, 6201-001, Covilh\~{a}, Portugal}
\affil[2]{NOVA LINCS, Universidade da Beira Interior, Covilh\~{a}, Portugal.}
\affil[1]{Email id: saeid.alirezazadeh@gmail.com}
\affil[2]{Email id: lfbaa@di.ubi.pt}
\date{}

\providecommand{\keywords}[1]{\textbf{\textit{Keywords---}} #1}
\begin{document}

\maketitle

\begin{abstract}
A scheduling method in a robotic network cloud system with minimal makespan is beneficial as the system can complete all the tasks assigned to it in the fastest way. Robotic network cloud systems can be translated into graphs where nodes represent hardware with independent computing power and edges represent data transmissions between nodes. Time window constraints on tasks are a natural way to order tasks. The makespan is the maximum amount of time between when the first node to receive a task starts executing its first scheduled task and when all nodes have completed their last scheduled task. Load balancing allocation and scheduling ensures that the time between when the first node completes its scheduled tasks and when all other nodes complete their scheduled tasks is as short as possible. We propose a grid of all tasks to ensure that the time window constraints for tasks are met. We propose grid of all tasks balancing algorithm for distributing and scheduling tasks with minimum makespan. We theoretically prove the correctness of the proposed algorithm and present simulations illustrating the obtained results.
\end{abstract}

\keywords{Clouds, dynamic scheduling, task classifier, task scheduling, virtual machines (VMs), load balancing, makespan.}

%

\section{Introduction}
Roboticsystems are used in different aspects of human life, such as domestic
, industrial and manufacturing
, military
among other
. Tasks often are beyond the capability of a single robot. Then, a natural choice is to use several robots instead of a single robot such that they cooperatively perform the task. For instance, transporting big objects and surveillance of a large area, are tasks where the performance can be improved by the use of several robots instead of a single robot. Such a system is called a robotic network. 

The use of robotic networks also has limitations. For example, the capacity of a robotic network is higher than a single robot, but it is bounded by the collective capacity of all robots~\cite{Hu:2012}. To handle this limitation we may consider increasing the number of robots to increase the capacity, but in this case, we simultaneously increase the complexity of the system. Another limitation is most of the tasks related to human-robot interaction, such as object, face, and speech recognition are computationally demanding tasks. 

Cloud robotics can be used to handle some of the computational limitations of robots. It takes advantage of the internet and of cloud infrastructure for assigning computation and also to perform real- time sharing of large data~\cite{kehoe:2015}. An important aspect of a cloud-based robotic system is deciding whether to upload a newly arrived task to the cloud, processing it on a server (fog computing~\cite{bonomi:2012}) or executing it on any of the robots (edge computing~\cite{shi:2016}), the so-called, allocation problem. After solving the allocation problem, the result will be a set of tasks that should be performed by each processing unit. The scheduling problem deals with the problem of ordering the set of tasks, taking into account the priority of the tasks, their time constraints, and the precedence order between tasks, to answer the question of which task should be executed first from the set of tasks assigned to each processing unit. Restricting the scheduling problem to the cloud translates the problem into the decision of how to assign tasks to different virtual machines in a cloud. One of the goals of this work is to find a way to do the scheduling such that the makespan is as small as possible. The makespan is the maximum amount of time between when the first node to receive a task starts executing its first scheduled task and when all nodes have completed their last scheduled task. This is important if we wish to have the tasks completed as quickly as possible. This requirement is common for anything that has to do with human-robot interaction, to avoid making a human wait for too long to get feedback from a robot. Another goal is to balance the loads of the processing units, which ensures that all processing units perform certain tasks and that the system uses its maximum computational capacity to complete all the assigned tasks, i.e., the processing units have the least $\Idle$ time and most of the consumed energy is used to perform tasks instead of being $\Idle$.

Tasks may have an order of execution, i.e. a task may be executed after another task has been completed or a task may have a limited time for execution, e.g. in manufacturing a product should be assembled before packaging or in disaster management tasks should normally be executed in a limited time. These are called time-window constraints.In this paper, we define the grid of all tasks, which maintains the order of tasks defined by the time-window constraints. 

Solving the allocation problem gives us a partitioning of the set of tasks, where the tasks in each part should be performed by one of the processing units. However, this does not generally give us the order in which the tasks in each part should be executed. Thus, we need to solve the scheduling problem, which requires information about the priority of the tasks, the time window, precedence order and so on. In this paper, we focus on minimizing the makespan while balancing the load of scheduled tasks for all nodes. Most of the existing results take into account the constraints induced by the problem over tasks and processing units. However, such constraints can only be determined for some specific problems, and they do not have general patterns, as they may vary from one problem to another. We present simulations that are randomly generated to discard possible hidden patterns that might emerge in experiments, allowing us to cover a wide range of settings.

The paper is organised as follows. Section 2 reviews related work on task assignment and scheduling in robotic network cloud systems. Section 3 introduces some basic and general concepts on task relations. Section 4 is an overview of the proposed method. Section 5 introduces the key algorithm that is central to this work. In Section 6, we propose a single grid for all tasks to determine the order in which the tasks are executed. In Section 7, we provide the proof of optimality and correctness of the key algorithm. Section 8 describes the experimental methodology and discusses the results of the experiments. In Section 9, we analyze the complexities of our key algorithm with the state-of-the-art algorithms. Section 10 provides the scalability analysis of our key algorithm and of the state-of-the-art algorithms. And finally, in Section 11, we draw some conclusions and point to future lines of work.

\section{Related Work}
Let $T$ be a finite set of tasks that can be performed by a multi-robot system. Consider the case where the system is performing a subset of tasks, say $T_1$, when a new set of tasks, $T_2$, arrives. There are two types of task allocation:
\begin{itemize} 
\item If the set of all tasks, $T$, is known, and we are looking for the optimal performance from the system by allocating the set of all tasks, then we call this type of task allocation \textbf{static task allocation}.
\item If sequences of tasks arrive to be performed by the system at different time steps, and we are looking for the optimal performance by dynamically allocating the newly arrived tasks, then we call this type of task allocation a \textbf{dynamic task allocation}.
\end{itemize}

In static task assignment, \cite{Lin:1995} studied static algorithm allocation for a multi-robot system without considering cloud infrastructure and communication times; \cite{Ours:2020} solved the allocation problem by minimizing memory and time for a single robot cloud system; and \cite{li:2018} approaches the assignment problem incompletely by considering only the minimum time and ignoring the memory parameter. Their method only minimizes the total execution time without fully considering communication, as shown in \cite{Ours:2020}. \cite{Ours:2021} solved the allocation problem by minimizing memory and time for a multi-robot cloud system.

Dynamic task assignment is divided into a centralized and distributed assignment, where in centralized methods, a central planning unit that has information about the entire environment and handles task assignment, \cite{Burkard:2012}, and in a distributed assignment, instead of a central unit, all tasks are distributed to all robots, and the robots decide which tasks to perform. For distributed assignment, the following five approaches are the mainly considered: \textbf{behavior-based}, where decision for considering a task by a robot is based on the problem features, \cite{Parker:1998}; \textbf{market-based}, where decision for considering a task by a robot is based on an auction-based mechanism, \cite{Alaa:2013}; \textbf{combinatorial optimization-based}, where decision for considering a task by a robot is made by transforming the problems into combinatorial optimization problems and using an appropriate existing technique to solve them, \cite{WANG:2020}; \textbf{evolutionary algorithm-based}, where decision for considering a task by a robot is made by using evolutionary operators on the space of solutions, \cite{Lane:2018}; \textbf{machine learning algorithm}, where a machine learning algorithm is used to find an optimal task assignment, \cite{Ding:2020}. For task assignment in a cloud robotic system, most studies only move heavy computational tasks to the cloud without considering the communication between robots and the cloud infrastructure, \cite{Slam:2015, Rapyuta:2013}.

Dynamic task assignment is widely studied. The most recent works are: \cite{Chen:2018} which proposed an algorithm to optimize the latency, energy consumption, and computational cost considering the cloud robotic architecture characteristics, task characteristics, and quality of service problems for cloud robotics; 
\cite{Zhang:2018} proposed a method to handle the deadline of a task and also minimize the total cost; \cite{Chen:2019} studied the task assignment under the assumption that the number of robots changes and the tasks are changeable; \cite{WANG:2020} studied the optimal task assignment for the case where two collaborative robot teams jointly perform some complex tasks, (they provide the optimization statement formulation of their model based on set-theoretic parameters); and \cite{Ours:2020h} translates optimal task scheduling into finding the maximum volume of a subspace of a hyperspace, (they studied compatibility of a node to perform a task, communication and communication instability, and capability of fog, cloud, and robots). For a set of newly arrived tasks, these studies mainly focused on methods to sequentially optimally assign each of the new tasks to a node of the architecture, ignoring the effects of the assigned tasks on the assignment of future sets of tasks; \cite{Ding:2020} has proposed a method to find an optimal solution for distributing tasks to servers that minimizes the cost to the user. In the proposed method, the task allocation problem is translated into a reinforcement learning problem where the reward function is the negative average user cost.

A recent paper~\cite{Yu:2021} described a mathematical model for load balancing and proposed two solutions, the Basic Load Balancing Algorithm (BLBA) and the Improved Load Balancing Algorithm (ILBA), which are based on a greedy strategy that measures the loads taking into account the interdependencies between the tasks and then uses an optimization model to minimize the mean square root of the loads. In the experimental part, we will compare our approach with these two methods.

The problems we address in this paper focus on the dynamic allocation problem. Our main goal is to define a method to assign the newly arrived tasks to different computational processing units in such a way that the overall makespan is minimal, while the times for completing the tasks are close to each other. 

Note that the actual task performance is always done by robots. However, robots take advantage of the Internet and cloud infrastructure for assigning computations and sharing data in real- time. This enables faster processing of a task by executing multiple subtasks (algorithms) in the cloud, the results of which are used together to execute the task by a robot. Using static allocation, we find out which tasks should be offloaded to the cloud. Then the problem of load balancing and minimizing the makespan becomes a problem of scheduling in the cloud.

The load balancing problem we addressed has similarities with results related to load balancing for multi-virtual machines (multi-VM) and multi-containers, and also has similarities with studies on the class of online scheduling problems on parallel machines.

In a recent work on multi-VM and multi-container, \cite{Patra:2020}, a method is developed that uses Balls in Bins, a greedy algorithm, but with a local search that checks if there are more tasks assigned to the neighboring physical servers in the cloud to assign them a new task or not. In our method, we do not consider the advantages and benefits of containerization instead of using virtual machines. Instead, we provide a solution for existing machines instead of creating new machines, i.e., if a user wants to run multiple jobs, it is not necessary to create a new machine for a particular new job, but we use the existing ones created for other jobs. We prove that our algorithm provides the optimal assignment that minimizes the makespan and balances loads, regardless of the constraints of the problems. 

In the class of online scheduling problems on parallel machines, we first assign tasks to the machines based on a policy (which can be a stochastic process or a greedy process) and then apply a policy for scheduling on each machine separately. The class of online scheduling on parallel machines is slightly different from the problem we are dealing with. In our problem, load balancing is centralized, and a central entity assigns tasks to the virtual machines. For this type of problem, it is necessary to apply our method twice, first to all machines and then to the processors of each machine. In \cite{Buchem:2021}, a stochastic online scheduling is used on uniform machines, and then the greedy algorithm was used as policies for load balancing.

Table~\ref{tab0} lists the notation used in the paper.
\begin{table}[!t]
\caption{Table of notation.}
\begin{center}
\begin{tabular}{p{0.12\linewidth}|p{0.8\linewidth}}
Notation&Description\\
\hline
$(a,b)$&The time window with earliest start time $a$ and latest finish time $b$\\
$\GT (a,b)$&The grid of all tasks with earliest start time $a$ and latest finish time $b$\\
$MNST$&Maximum number of tasks that are not $\Idle$ scheduled within all nodes\\
$N$&A node of the architecture\\
$\mathcal{N}$&The set of all nodes\\
$N_x$&The $x$-th node of the architecture\\
$NT_N$&The maximum number of tasks that node $N$ can perform simultaneously\\
$NT$&The sum of $NT_N$ for all $N$\\
$\mathbf{P}$& The set of all processing units\\
$STS(N)$&The list of all scheduled tasks for node $N$\\
$STS(\mathcal{N})$&The list of all scheduled tasks for all nodes\\
$T$&An individual task\\
$\mathbf{T}$& The set of newly arrived tasks\\
$\mathcal{T}$& The set of all tasks\\
$T^i_j(N)$&The $i$-th task in the list of scheduled tasks for the $j$-th processors of node $N$\\
$t_e^N(T)$&The average time for node $N$ from the start to the completion of task $T$\\
$t_s^N(T)$&The actual start time of the task $T$ by the node $N$
\end{tabular}
\end{center}
\label{tab0}
\end{table}
\section{Method Overview}
\begin{figure}[!tb]\centering
\includegraphics[width=0.5\linewidth]{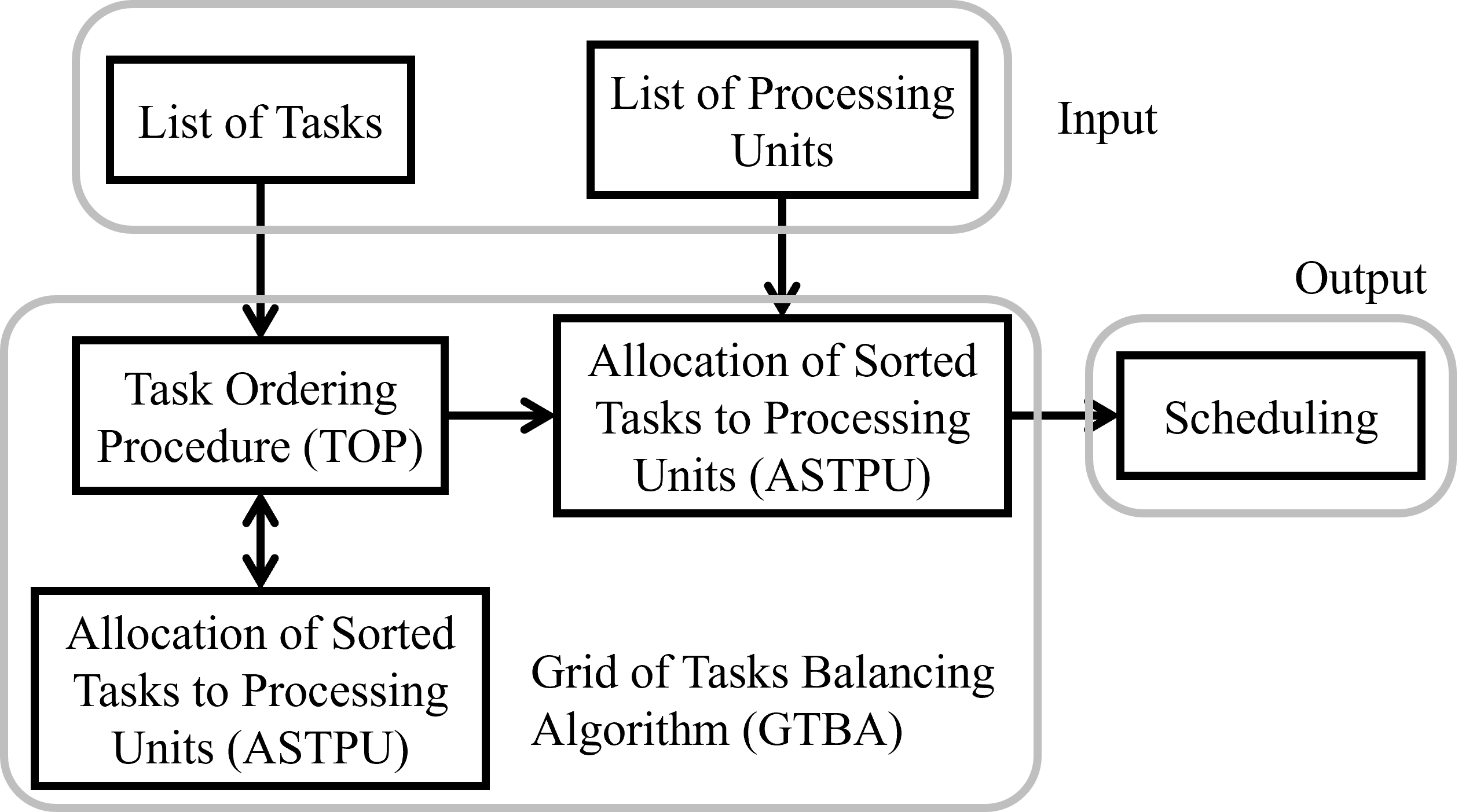}
\caption{The intuition of the proposed method to minimize the makespan and balance the loads.}
\label{fig0}
\end{figure}

We propose a new method to minimize the makespan and balance the loads, see Figure \ref{fig0}. A list of tasks arrives in the system with a list of processing units that may have some scheduled tasks. Use the grid of all tasks balancing algorithm (GTBA) to assign each task to a processing unit. In the GTBA block, the allocation of sorted tasks to processing units (ASTPU) sorts the tasks and processing units based on the execution time of the tasks and the completion time of the scheduled tasks on the processing units and determines which task should be scheduled for which processing units. The task ordering procedure (TOP) uses the ASTPU and the time window constraints of the tasks to create the grid of all tasks. Then, the ASTPU is used for the grid of all tasks with the list of processing units to identify the tasks to be scheduled.

\section{Preliminaries}

We begin with a description of the mathematical tools used in this paper. All tasks are considered to have time windows, that is, the interval $[a,b]$, where $a$ is the earliest start time and $b$ is the latest finish time (or deadline). Moreover, each task $T$ has an average completion time when executed by node $N$, denoted by $t_e^N(T)$, that is the average time for node $N$ from the start to the completion of task $T$.

If the earliest start time of a task is not specified, the time window is considered as the interval $[0,b]$ and if the deadline is not specified, the time window is considered as the interval $[a,\infty]$.

Next, we describe the expressions for the relations between the time windows of the tasks. Let $T_1$ and $T_2$ be two tasks with time windows $[a_1,b_1]$ and $[a_2,b_2]$, respectively, then we write:
\begin{enumerate}[label=\textbf{R.\arabic*}]
\item\label{1} $T_1< T_2$, when the task $T_1$ has to be performed before the task $T_2$ starts, i.e., $a_2>b_1$.
\item\label{2} $T_1\wedge T_2$, when the task $T_2$ has to be performed immediately after the task $T_1$, i.e., $a_2=b_1$.
\item\label{3} $T_1\Rightarrow T_2$, when the task $T_1$ has to be started after the task $T_2$ starts and also the task $T_1$ is needed to be finished before the task $T_2$ finishes, i.e., $a_1>a_2$ and $b_1< b_2$.
\item\label{4} $T_1\vdash T_2$, when the tasks $T_1$ and $T_2$ have to be both started at the same time, but the task $T_2$ finishes after the task $T_1$ is finished, i.e., $a_1=a_2$ and $b_1<b_2$.
\item\label{5} $T_1\dashv T_2$, when the tasks $T_1$ and $T_2$ have to be both finished at the same time, but the task $T_1$ starts before the task $T_2$ starts, i.e., $a_1<a_2$ and $b_1= b_2$.
\item\label{6} $T_1=T_2$, when the tasks $T_1$ and $T_2$ have to be both started and finished at the same time, i.e., $a_1=a_2$ and $b_1=b_2$.
\item\label{7} $T_1\vee T_2$, when the task $T_2$ starts after the task $T_1$ starts and also the task $T_2$ finishes after the task $T_1$ finishes, i.e., $a_1< a_2<b_1<b_2$.
\item\label{8} $T_1\wr T_2$ if none of the above are needed to be considered as necessary conditions for the tasks $T_1$ and $T_2$.
\end{enumerate}
With the preceding notation we can also describe priorities of performing tasks in detail. It is easy to observe that for tasks $T_1,T_2$, and $T_3$, $T_1\wedge T_2$ and $T_1\wedge T_3$ implies one of the following holds: $T_2\vdash T_3$ or $T_3\vdash T_2$ or $T_2=T_3$. One can classify tasks with respect to the binary relation $\sim\in\{<,\wedge,\Rightarrow,\vdash,\dashv,=,\vee,\wr\}$ as $[T]_{\sim}=\{T'\mid T'\in\mathcal{T} \text{ and } T\sim T'\}$, 
where $\mathcal{T}$ is the set of all tasks. Note that, by definition, for $\sim_1,\sim_2\in\{<,\wedge,\Rightarrow,\vdash,\dashv,=,\vee,\wr\}$, 
with $\sim_1\neq\sim_2$, we have $[T]_{\sim_1}\cap[T]_{\sim_2}=\emptyset$, and 
\begin{equation}\label{eq:partition}
\bigcup_{\sim\in\{<,\wedge,\Rightarrow,\vdash,\dashv,=,\vee,\wr\}}[T]_{\sim}=\mathcal{T}.
\end{equation}
Hence, the set of non-empty classes $[T]_{\sim}(\neq\emptyset)$ can be considered as a partition of $\mathcal{T}$ which is the set of all tasks. 

Note that, for every node, scheduled tasks are performed according to task priorities. For a node $N$, let $NT_N$ be the maximum number of tasks that the node $N$ can perform simultaneously. Denote by
$$STS(N)=\left((T^i_1(N),\ldots,T^i_{NT_N}(N))\mid T^i_1\in\mathcal{T}\cup\{\Idle\}\right)_{i\in\mathbb{N}}$$
the list of all scheduled tasks on the node $N$, where the index $i\in\mathbb{N}$ is used for the scheduling order. We call the sequence of scheduled tasks in each component of $STS(N)$, a stream. If $NT_N=1$, then node $N$ is a single task node and if $NT_N>1$, then node $N$ is a multi-task node. For a node $N$ and $j\in\{1,\ldots,NT_N\}$, we call the $j$-th component of $STS(N)$,
$$STS(N)\mid_{j-\text{th component}}=(T_j^i(N))_{i\in\mathbb{N}}$$ 
the $j$-th stream scheduled tasks of the node $N$. The $\Idle$ is used to identify the last tasks, i.e,  if $T_j^k(N)=\Idle$, then for all $i$, with $i\geq k$, $T_j^i(N)=\Idle$. The index $k$ identifies the final task, and the index $i$ which we consider $i \geq k$ is used to say all the other tasks are $\Idle$. This means that there are no more scheduled tasks for the $j$-th stream. Define $STS(\mathcal{N})=\{STS(N)\mid N\in\mathcal{N}\}$, where $\mathcal{N}$ is the set of all nodes. The time-frame of the $j$-th stream scheduled tasks of the node $N$ can be evaluated as an ordered sequence of elements as follows:
$$(0,t_e^N(T_j^1))\times\Bigg(t_e^N(T_j^1)+\left(\sum_{k=1}^it_e^N(T_j^k),\sum_{k=1}^{i+1}t_e^N(T_j^k)\right)\Bigg)_{i\in\mathbb{N}}.$$
If $T_j^k=\Idle$, then $t_e^N(T_j^k)=0$.

We can find the maximum number of scheduled tasks within all nodes, denoted by $MNST$, that is the maximum length of tasks scheduled at streams of nodets which are not $\Idle$, i.e., 
\begin{align*}
MNST=\max&\Big\{k\mid~T_j^k(N)\neq\Idle, T_j^{k+1}(N)=\Idle, N\in\mathcal{N},\\
&~~\text{ and }~ STS(N)\mid_{j-\text{th component}}=(T_j^i(N))_{i\in\mathbb{N}}\Big\}.
\end{align*}
Then, the time frame can be viewed as a grid $MNST\times\sum_{N\in\mathcal{N}}NT_N$, where the value of the $(i,j)$ position is equal to $(a,b)$, and there is some positive integer $k$ such that 
$$\sum_{x=1}^{k-1}NT_{N_x}< i\leq \sum_{x=1}^{k}NT_{N_x}$$ 
and $(a,b)$ is the $j$-th component of the time frame of the $y$-th stream of the node $N_k$, where $y=i-\sum_{x=1}^{k-1}NT_{N_x}$. After creating the grid, we can observe that there is, by construction, a one-to-one correspondence between the set of scheduled tasks of all the nodes and each component of the time frame. 

We are assuming that, for the existing scheduled tasks of all nodes, the relationships~\ref{1}-\ref{8} are satisfied. 

Now, assume that a task $T$ is newly introduced and needs to be allocated to a node to be performed. We first sort the grid in decreasing order by considering first the values of the last row and second the number of their scheduled tasks, if their values are equal.

Tasks co-occurring means: if a task $T$ arrives, then so do all the tasks in
$$\bigcup_{\sim\in\{<,\wedge,\Rightarrow,\vdash,\dashv,=,\vee\}}[T]_{\sim}\subseteq\mathcal{T}.$$
Note that, here we drop and do not consider the relation $\wr$, because positions of tasks in class $\wr$ will not affect the positions of other tasks, in other class, in the grid of all tasks.

We are assuming that co-occurring tasks do not appear. This means that a new task $T$ can be performed independently from the occurrence of the set of tasks in 
$$\bT=\bigcup_{\sim\in\{<,\wedge,\Rightarrow,\vdash,\dashv,=,\vee\}}[T]_{\sim}\subseteq\mathcal{T}.$$

\begin{remark}\label{rem1}
The following holds concerning the classes of tasks that are performed by nodes:
\begin{enumerate}[label=\textbf{S.\arabic*}]
\item\label{1s} For a task $T$, the set of all tasks in classes $[T]_=, [T]_{\vee}, [T]_{\dashv}, [T]_{\vdash}$, and $[T]_{\Rightarrow}$ need to be performed with different streams in nodes, because each stream can only perform a single task at a time and tasks within each of these classes need to be performed in parallel.
\item\label{2s} For two tasks $T_1$ and $T_2$, if $[T_1]_=\cap[T_2]_=\neq\emptyset$, then $[T_1]_==[T_2]_=$.
\item\label{3s} Let, for two tasks $T_1$ and $T_2$, the equality $[T_1]_=\cap[T_2]_==\emptyset$ hold. If there are tasks $T'_1,T'_2\in[T_1]_=$ and $T''_1,T''_2\in[T_2]_=$ such that for some binary relations $\sim_1,\sim_2\in\{<,\wedge,\Rightarrow,\vdash,\dashv,=,\vee\}$, the relationships $T'_1\sim_1 T''_1$ and $T'_2\sim_2T''_2$ hold, then $\sim_1=\sim_2$.
\item\label{4s} Let, for two tasks $T_1$ and $T_2$, the equality $[T_1]_=\cap[T_2]_==\emptyset$ hold. If there are tasks $T\in[T_1]_=$ and $T'\in[T_2]_=$ such that for a binary relation $\sim\in\{<,\wedge,\Rightarrow,\vdash,\dashv,=,\vee\}$, the relationship $T\sim T'$ holds, then for all $S\in[T_1]_=$ and $S'\in[T_2]_=$, the relationship $S\sim S'$ holds.
\item\label{5s} For two tasks $T_1$ and $T_2$, if the conditions in \ref{2s} and \ref{3s} are satisfied, where $\sim,\sim_1,\sim_2\in\{\Rightarrow,\vdash,\dashv,=,\vee\}$, then all tasks in $[T_1]_=\cup[T_2]_=$ need to be performed with different streams in nodes as in \ref{1s}.
\item\label{6s} For two tasks $T_1$ and $T_2$, if the conditions in \ref{2s} and \ref{3s} are satisfied, where $\sim,\sim_1,\sim_2\in\{<,\wedge\}$, then the set of all tasks in $[T_1]_=$ and also the set of all tasks in $[T_2]_=$ need to be performed with different streams in nodes as in \ref{1s}, but each task from $[T_1]_=$ and each task from $[T_2]_=$ can be performed with the same stream in nodes.
\end{enumerate}
\end{remark}
\begin{definition}\label{def1}
By Remark~\ref{rem1}, one can define the relationships $\mathrel{parallel}$ and $\mathrel{serial}$ between tasks. For two tasks $T_1$ and $T_2$, we say in view of Remark~\ref{rem1}, that $T_1$ and $T_2$ are parallel, denote by $T_1\mathrel{parallel}T_2$ if and only if $T_1$ and $T_2$ must be performed in different streams in nodes, and that $T_1$ and $T_2$ are serial, denote by $T_1\mathrel{serial}T_2$ if and only if $T_1$ and $T_2$ can be performed in the same stream in nodes.
\end{definition}
\section{Key Method (Allocation of Sorted Tasks to Processing Units)}

We first describe the Allocation of Sorted Tasks to Processing Units (ASTPU) that plays the key role in the paper. The proof of the correctness and optimality of the algorithm is given as the existence of a proof of the Proposition~\ref{prop:prop1} and Corollary~\ref{cor:cor1}. The ASTPU method is as follows:
\begin{algorithm}
\caption{Pseudocode of allocation of sorted tasks to processing units (ASTPU)}\label{algpseud}
\begin{algorithmic}[1]
\Require Set of newly arrived tasks $\mathbf{T}$ and set of scheduled tasks for processing units $\mathbf{P}$.
\Ensure The mapping $M:\mathbf{T}\rightarrow\mathbf{P}$ from the set of newly arrived tasks to the set of processing units.
	\While{$\mathbf{T}\neq\emptyset$} \Comment{There is at least one task to be assigned to a processor.}
		\State Sort the sets $\mathbf{T}$ in descending order \Comment{With respect to their execution time.}
		\State Sort the set $\mathbf{P}$ in ascending order \Comment{With respect to the completion time of all their scheduled tasks.}
		\State $M(T)=P$. \Comment{Assign the first task $T$ from the sorted set of tasks $\mathbf{T}$ to the first processing unit $P$ from the sorted set of processing units $\mathbf{P}$.}
		\State Update the set of tasks $\mathbf{T}=\mathbf{T}\setminus\{T\}$.
		\State Update the set of scheduled tasks for the processing units. \Comment{All processing units have their initial scheduled tasks except for processing unit $P$, which has a new task $T$ assigned and added to its scheduled tasks.}
	\EndWhile	
\Return The mapping $M$.
\end{algorithmic}
\end{algorithm}

\section{Task Ordering Procedure}
For a finite set of nodes $\mathcal{N}=\{N_1,\ldots,N_n\}$ recall that $NT_N$ is the number of tasks that the node $N$ can perform simultaneously, and let 
$NT=\sum_{N\in\mathcal{N}}NT_N$.

Let $\mathcal{T}=\{T_1,\ldots,T_m\}$ be a finite set of tasks which also include the forced $\Idle$ task, that corresponds to waiting for a specific amount of time. Let $\bT$ be the set of tasks newly arrived to the system that needs to be scheduled to nodes. The grid of all tasks can be created as follows: (In the following, if the number of rows of the grid is larger than $NT$, then the robotic system $\mathcal{N}$ is not suitable to perform the set of tasks $\mathcal{T}$)
\begin{enumerate}[label=\textbf{A.\arabic*}]
\item\label{0a} Find the set of all tasks in $\mathcal{T}\setminus\bT$.
\item\label{1a} Find the minimal task, $T$, with respect to the relation $<$ within all tasks from $\bT$.
\item\label{2a} Find sets of all distinct classes of tasks in view of Remark~\ref{rem1}, and consider those containing $T$. 
\item\label{3a} Identify the $\mathrel{parallel}$ classes of tasks.
\item\label{4a} Place the task $T$ time window in the first row of the grid, and other $\mathrel{parallel}$ tasks on the following rows consecutively.
\item\label{5a} Update the list of tasks by removing all the tasks that are already placed in the grid form $\bT$, and continue if the updated set is non-empty. If it is empty, go to step~\ref{10a}.
\item\label{6a} Apply the rules~\ref{1a}-\ref{3a} to the updated set of tasks.
\item\label{7a} Find in the grid the interval $(a,b)$, where $a$ is the smallest of the earliest start times and $b$ is the largest of the deadlines, in the existing tasks in the grid. 
\item\label{8a} Place the new task $T$ (the minimal task of the updated set) time window in the existing row with the minimum deadline, and other $\mathrel{parallel}$ tasks on the following rows consecutively (with minimum deadlines). Here we use the ASTPU method, where each row represents a processing unit. If the number of tasks in the updated set of tasks is larger than the number of rows in the grid, then we need to add new rows with forced $\Idle$ task with $(a,b)$ waiting time as in \ref{7a} and place the remainder of tasks according to the ASTPU method~\ref{algpseud}.
\item\label{9a} Return to step~\ref{5a}.
\item\label{10a} Find $(a,b)$ as in step~\ref{7a}.
\item\label{11a} Place all the tasks in step~\ref{0a} in the grid with possible consideration of additional rows up to $NT$ rows\footnote{New rows will be added only if placing new task increases the variance. The variance is evaluated from considering time windows of tasks in $\mathcal{T}$ by considering the forced $\Idle$ tasks in the grid.} in the grid. The placement is made by using the ASTPU method~\ref{algpseud}, and the tasks may replace the forced $\Idle$ tasks, if possible\footnote{If the time window of a task is smaller than the time window of the forced $\Idle$ task.}.
\end{enumerate}

The grid of all tasks is an optimal task ordering in which the tasks satisfy conditions~\ref{1}-\ref{8}, the minimum number of nodes are used to perform the tasks, and the total $\Idle$-time by streams of nodes in use is minimal.

The grid of all tasks can be considered as a single (black-box) task with the time window $(a,b)$ as in step~\ref{7a}.

Let $k$ be the number of rows of the grid of all tasks and $NT$ be the total number of streams. To have a minimum total time for $\Idle$, we must have $k\mid NT$ ($k$ divides $NT$), because since we are assigning a new task using the pigeonhole principle, there are some streams (exactly equal to the remainder of $NT$ by $k$) that are always $\Idle$.

Since the scheduled tasks for the nodes must satisfy conditions~\ref{1}-\ref{8}, the scheduled tasks can be viewed as a union of grids of all tasks, where some of the tasks can be replaced by a forced $\Idle$ task.

\begin{remark}\label{rem2}
Let the number of tasks in the class $=$ be equal to $k$ (which is found by expanding the relations in Remark~\ref{rem1}). Note that in theory, each of the $k$ tasks can be considered to be allocated to different streams of nodes, but the nodes must have exactly the same processing power. In reality, the same performance may not be achieved due to unexpected delays. Therefore, we assume that the compatible robotic system is such that there is at least one node $N$ such that $NT_N > k$. If there is no such node, we say that the robotic system is not compatible to perform the tasks. From now on, the tasks of class $=$ can only be allocated to a node $N$ with $NT_N > k$.
\end{remark}
\subsection*{Grid of all Tasks Balancing Algorithm (GTBA)}
Denote by $\GT(a,b)$ the grid of all tasks where $(a,b)$ is defined as in step~\ref{7a}. Let $\GT_1(a_1,b_1)$ and $\GT_2(a_2,b_2)$ be two grids of all tasks with their time windows. We say that $\GT_1(a_1,b_1)$ is smaller than $\GT_2(a_2,b_2)$, and denote it by $\GT_1(a_1,b_1)<\GT_2(a_2,b_2)$, whenever $a_1\leq a_2$. 

For a task $T$ with time window $(c,d)$ and $\GT(a,b)$ we say that $\GT(a,b)$ accepts the task $T$, if the time window of task $T$ in $\GT(a,b)$ contains $(c,d)$. 

Consider that a new task $T\in\mathcal{T}$ is introduced to the system to be allocated to a stream with time window $(c,d)$. Define $\GT(a,b)\mid_T$ to be the grid of all tasks with time window $(a,b)$, if the time window of task $T$ is equal to $(c,d)$ and all the tasks except $T$ are replaced with a forced $\Idle$ task. 

We are using the ASTPU method~\ref{algpseud} for allocation of tasks to processing units as follows: For the task $T$, find $t_s^N(T)$, the actual start time of the task $T$ on the node $N$, and then, under the setting of Remark~\ref{rem2}:
\begin{itemize}
\item For all the existing grids, use the ASTPU method to allocate task $T$ to the stream of the smallest grid $\GT(a,b)$ that accepts the task $T$ with time window $t_s^N(T)+(c,d)$, such that the forced $\Idle$ task is in the position of the task $T$ in the grid $\GT(a,b)$.
\item If the preceding does not hold\footnote{Either none of the existing grids accepts the task $T$ or there are some grids that accept $T$ but in those grids in the time window of task $T$, a non-$\Idle$ task is already scheduled.}, then create $\GT(a,b)\mid_T$ and allocate all the tasks in $\GT(a,b)\mid_T$ to the first $k$ streams as follows. Allocate the smallest row of the grid $\GT(a,b)\mid_T$ to the stream with the smallest finishing time using the ASTPU method. Then recursively allocate the next smallest row of $\GT(a,b)$ to the next stream with the smallest finishing time\footnote{To preserve the order of tasks in $\GT(a,b)\mid_T$, for allocating $\GT(a,b)\mid_T$ to $k$ streams we may need to add some forced $\Idle$ tasks to change the start time of the first tasks in some rows of $\GT(a,b)\mid_T$, i.e., put some delays in order to preserve the shape of the grid.} using the ASTPU method.

Note that, by Remark~\ref{rem2}, tasks in class $=$ must be allocated to the same node.
\end{itemize}
The preceding task allocation preserves the order of tasks induced by conditions~\ref{1}-\ref{8}, is able to identify an optimal number of streams for the in-use robotic system, provides a smaller number of $\Idle$ streams, the position of new tasks can be easily identified, and total time for performing all the tasks is minimized. 

Moreover, Remark~\ref{rem2} implies that the set of tasks that need to be performed simultaneously are necessarily allocated to streams of a single node. Now, by considering the set of newly arrived tasks as a grid of all tasks and applying the ASTPU to the streams of the grid considering the order of the grids, we can simultaneously solve the allocation and the scheduling problem with balanced loads and minimized makespan. Since by construction of the grid of all tasks, the time constraints of the newly arrived tasks are satisfied, we can avoid testing the time windows constraints and schedule only the tasks.

\section{Proof of optimality}
The following proposition and its corollary show that the ASTPU method~\ref{algpseud} is an optimal method for the task allocation in the sense that all processing units can complete their tasks almost at the same time. We show that assigning tasks to processors in a different way than the ASTPU method~\ref{algpseud} increases the variance meaning that the loads will be less balanced.
\begin{proposition}{(ASTPU method)}\label{prop:prop1}
Let $A=\{a_1, \ldots, a_n\}$ be a list of positive real numbers and $m\geq 1$ be an integer. Assume that we are randomly removing one of $a_i$'s and putting it in exactly one of the $m$ places and do the same process untill the set $A$ is empty. Let $X_j$, for $j=1,\ldots,m$, be random variables defined as the sum of $a_i$'s in the place $j$. Then, the ASTPU algorithm~\ref{algpseud} minimizes the variance of $X_i$'s.

Furthermore, for the case of $A=\{a_1,\ldots,a_n\}$ and $m\geq n$, the minimal variance can be obtained by putting exactly one of the $a_i$'s in each place.
\end{proposition}
\begin{proof}
We first show that: if $A=\{a_1,\ldots,a_n\}$ and $m\geq n$, then the minimal variance can be obtained by putting exactly one of the $a_i$'s in each place. In this case, without loss of generality we can assume that $X_i=a_i$ for $i=1,\ldots,n$ and $X_i=0$ for $i=n+1,\ldots,m$, then $\mu(X)=\frac{a_1+\ldots+a_n}{m}$ and 
$$Var(X)=\frac{1}{m}\sum_{i=1}^n(a_i-\mu)^2+\frac{1}{m}\sum_{i=n+1}^m(\mu)^2.$$
Now, let $Y$ be a random variable obtained from $X$ by substituting exactly one of the non-zero values of $X$ with zero and add the same amount to exactly one of the other non-zero values of $X$. Then we compare their variance and show that the variance of $Y$ is larger than the variance of $X$. Comparisons with variances for other scenarios of allocations can be evaluated simply by iterating the same arguments. Without loss of generality, we can assume that $Y_1=X_1+X_2$, $Y_n=0$ and $Y_i=X_{i+1}$ for $i=2,\ldots, n-1$. Then 
$$Var(Y)=\frac{1}{m}(a_1+a_2-\mu)^2+\frac{1}{m}\sum_{i=2}^{n-1}(a_i-\mu)^2+\frac{1}{m}\mu^2+\frac{1}{m}\sum_{i=n+1}^m(\mu)^2.$$
Hence,
\begin{align*}
Var(Y)-Var(X)=&\frac{1}{m}(a_1+a_2-\mu)^2+\frac{1}{m}\mu^2-\frac{1}{m}(a_1-\mu)^2-\frac{1}{m}(a_2-\mu)^2\\
=&\frac{1}{m}(2a_1a_2)>0,\\
\end{align*}
which implies that $Var(Y)>Var(X)$. 
Now, assume that $A=\{a_1,\ldots,a_n\}$ and $m\leq n$. Let $\mu=\frac{a_1+\ldots+a_n}{m}$. By the pigeonhole principle, see~\cite{Herstein:1964}, there are some places with at least two values from $A$. The ASTPU method~\ref{algpseud} is based on recursive allocation and sorting, see Figure~\ref{fi2}.
\begin{figure}[!t]\centering
\includegraphics[width=0.5\linewidth]{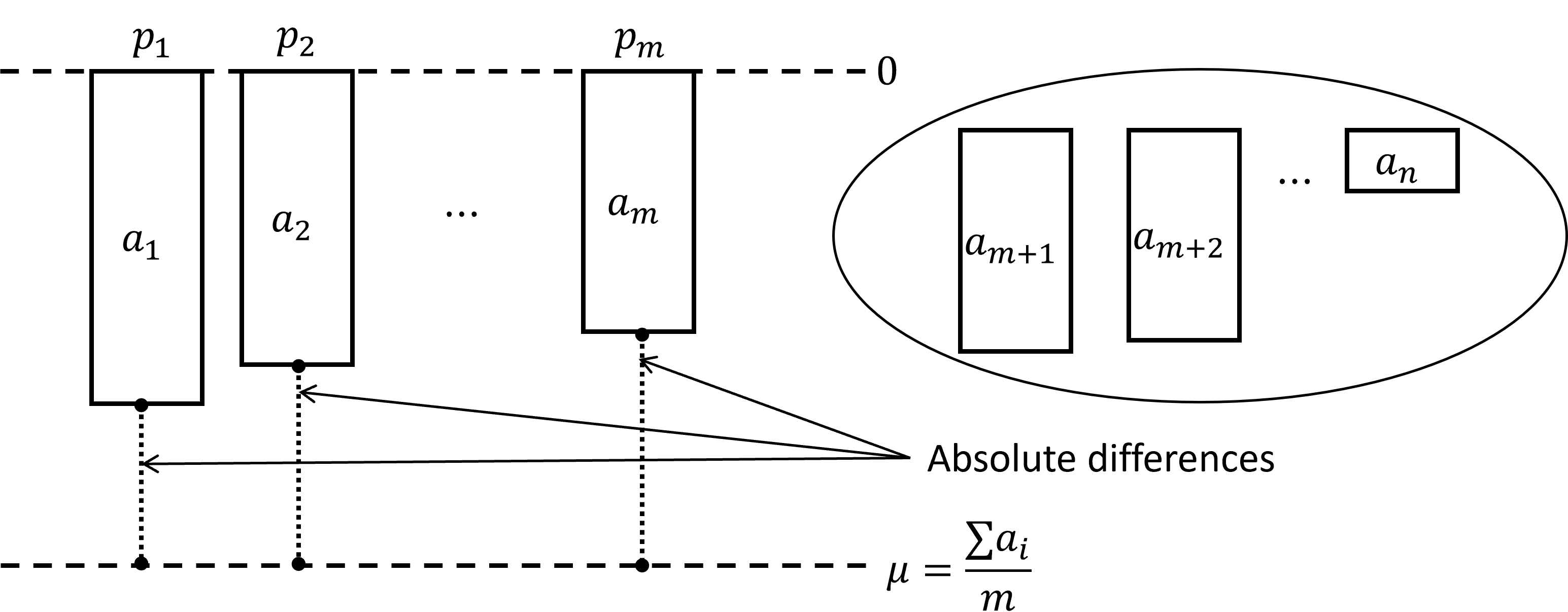}
\caption{Creating columns with blocks on top of each other such that the total square differences between the length of columns and the $\mu$ line is minimum.}
\label{fi2}
\end{figure}
We first associate randomly $m$ values from $A$ to $X_i$'s. For simplicity we are assuming that $X_i=a_i$ for $i=1,\ldots,m$. Now, we want to add $a_{m+1}$ to either $X_i$ or $X_j$ where we only know that $X_i<X_j$. We consider the addition to both and compare the effect on the variances. Note that, the following terms have appeared in both variances
$$\frac{1}{m}\sum_{\scriptsize{\begin{array}{c}k=1,\\ k\neq i,j\end{array}}}^m(X_k-\mu)^2.$$
The only differences are the following terms
\begin{align}
X_i\rightarrow X_i+a_{m+1}:&\quad (\mu-X_i-a_{m+1})^2+(\mu-X_j)^2,\label{eq:adi}\\
X_j\rightarrow X_j+a_{m+1}:&\quad (\mu-X_i)^2+(\mu-X_j-a_{m+1})^2,\label{eq:adj}
\end{align}
where $X_i\rightarrow X_i+a_{m+1}$ stands for the value $a_{m+1}$ which is added to $X_i$. Note that by adding $a_{m+1}$ to either $X_i$ or $X_j$, all terms $(X_k-\mu)^2$ for $k\neq i,j$ remain the same when calculating the variances in both cases. However, when we consider the addition of $a_{m+1}$ to $X_i$ in the variance, the term $(X_i-\mu)^2$ is replaced by $(X_i+a_{m+1}-\mu)^2$, while $(X_j-\mu)^2$ remains unchanged, and when we consider the addition of $a_{m+1}$ to $X_j$ in the variance, the term $(X_j-\mu)^2$ is replaced by $(X_j+a_{m+1}-\mu)^2$, while $(X_i-\mu)^2$ remains unchanged. Thus, for the two cases where we add the term $a_{m+1}$ to either $X_i$ or $X_j$ and compute the differences between their variances, the terms in \eqref{eq:adi} and \eqref{eq:adj} are the changed terms (differences) in the variances of both cases. Now, by expanding the variance and removing similarities, we obtain $-2(\mu-a_i)$ and $-2(\mu-a_j)$, respectively, from \eqref{eq:adi} and \eqref{eq:adj} as their differences. On the other hand, the assumption $X_i<X_j$ implies $X_i-\mu<X_j-\mu$. Hence, 
$$Var(X_i\rightarrow X_i+a_{m+1})<Var(X_j\rightarrow X_j+a_{m+1}).$$
And so, the value $a_{m+1}$ must be added to the 
\begin{equation}\label{eqtmp}
X_{k'}=\min\{X_i\mid i=1,\ldots,m\}.
\end{equation}

We show, by induction on the size of the list $A$, that the minimal variance appears whenever we use the first $m$ maximum values of the list for $X_i$'s, and then by the preceding process, once at a time, add the maximum of the remaining values to the corresponding place obtained in \eqref{eqtmp}, again once at a time. 

For the first step of the induction: we show that for $n=m+1$ the result holds. Assume that $X_i=a_i$ for $i=1,\ldots, m$ are chosen such that $a_1,\ldots,a_m$ are the first $m$ largest values of the list $A$, that is, $X_1\leq\cdots\leq X_m$. Let $Y_i=X_i$ for all $i$ except one, $Y_j=a_{m+1}$ where $a_{m+1}\leq a_m\leq a_j$. Now, we add $a_{m+1}$ and $a_j$ respectively to $X$ and $Y$. This is equivalent to saying we add $a_{m+1}$ to the minimum of $X_i$'s and add $a_{m+1}$ to other non-minimum $X_i$'s. Hence, by what we showed before, $Var(Y)\geq Var(X)$ and the equality holds only if $a_j=a_m$. 

The induction hypothesis is that for $n>m$ the result holds. It remains to show that the result holds for $n+1$. The result is immediate by using the allocation method for the sub-list $A'=A\setminus\{a\}$ with $a=\min A$ then applying the induction hypothesis together with the preceding arguments.
\end{proof}
\begin{corollary}\label{cor:cor1}
In Proposition~\ref{prop:prop1}, we can assume that some of the $m$ places have some initial fixed values. Then apply the method in the proof of the Proposition~\ref{prop:prop1} for the allocation of the numbers in the list of numbers under the assumption that some of the $m$ places have initial values. However, the method under the assumption of existing some initial values may not give the overall minimal variance comparing with the case without considering the initial values, but will give the optimal allocation such that the variance is minimal with respect to the existence of the initial values. 
\end{corollary}
\begin{proof}
Following the notation in the proof of Propositiom~\ref{prop:prop1}, we assume without loss of generality that $X_i\leq X_{i+1}$ for $i=1,\ldots,m$. We want to add $a_1$ and $a_2$ to the $X_k$'s such that the variance is minimal, where $a_1\leq a_2$. As we showed in the proof of Proposition~\ref{prop:prop1}, for any $a$, adding $a$ to any of the $X_k$'s has the minimum variance if and only if $a$ is added to the smallest $X_k$'s. This means that $a_1$ or $a_2$ should be added to the smallest $X_k$'s and then the other to the smallest $X_k$'s after the first addition is made. We compare the variances of the different scenarios. Note that the following terms appear in both variances $\frac{1}{m}\sum_{k=3}^m(X_k-\mu)^2$, the only differences are the following terms
\begin{align}
case~1:~&X_1\rightarrow X_1+a_1~\text{and}~X_2\rightarrow X_2+a_2:~ (\mu-X_1-a_1)^2+(\mu-X_2-a_2)^2,\label{eq:ad1}\\
case~2:~&X_1\rightarrow X_1+a_2~\text{and}~X_2\rightarrow X_2+a_1:~ (\mu-X_1-a_2)^2+(\mu-X_2-a_1)^2,\label{eq:ad2}
\end{align}
when calculating the variances in both cases. We calculate the differences between their variances. The terms in \eqref{eq:ad1} and \eqref{eq:ad2} are the changed terms (differences) in the variances of the two cases. Now, if we expand the variances and remove similarities, we get $2a_1X_1+2a_2X_2$ and $2a_1X_2+2a_2X_1$ from \eqref{eq:adi} and \eqref{eq:adj} as their differences. If $X_1=X_2$ and $a_1=a_2$, then both cases are identical and the variances remain the same. Suppose that $X_1 < X_2$, then in this case since $a_1\leq a_2$, we have $a_1(X_1-X_2)\geq a_2(X_1-X_2)$, which means $a_1X_1+a_2X_2\geq a_2X_1+a_1X_2$. Consequently, $Var (case 1)\geq Var (case 2)$. And if we assume that $a_1 < a_2$, then in this case, since $X_1\leq X_2$, we have $X_1(a_1-a_2)\geq X_2(a_1-a_2)$, which means $a_1X_1+a_2X_2\geq a_2X_1+a_1X_2$. Consequently, $Var(case 1)\geq Var(case 2)$.
\end{proof}
The method described in Proposition~\ref{prop:prop1} shows how the set of newly arrived tasks can be optimally distributed among the processing units, assuming that no tasks are initially assigned to the processing units. This method is extended by Corollary~\ref{cor:cor1}, which shows that it is optimal to assign tasks to the processing units if some processing units already have some tasks to complete.

\section{Experiments}
The experiments are performed on a HP Laptop 15-dw2xxx with Intel Core i5 10th generation with processor Intel(R) Core(TM) i5-1035G1 CPU @ 1.00GHz   1.19 GHz, RAM 16.0 GB (15.8 GB usable), 64-bit operating system, x64-based processor, and we used RStudio Version 1.4.1103 © 2009-2021 RStudio, PBC and the R version 4.0.3 (2020-10-10) copyright © 2020 The R Foundation for Statistical Computing Platform: x86\_64-w64-mingw32/x64 (64-bit).

We consider $10$ virtual machines, but the total number of tasks is not given. However, we assume that the execution time of each of the newly arrived tasks is randomly chosen by one of the virtual machines from the interval $[0,50]$ in milliseconds.

\begin{figure}[!t]\centering
\includegraphics[width=0.5\linewidth]{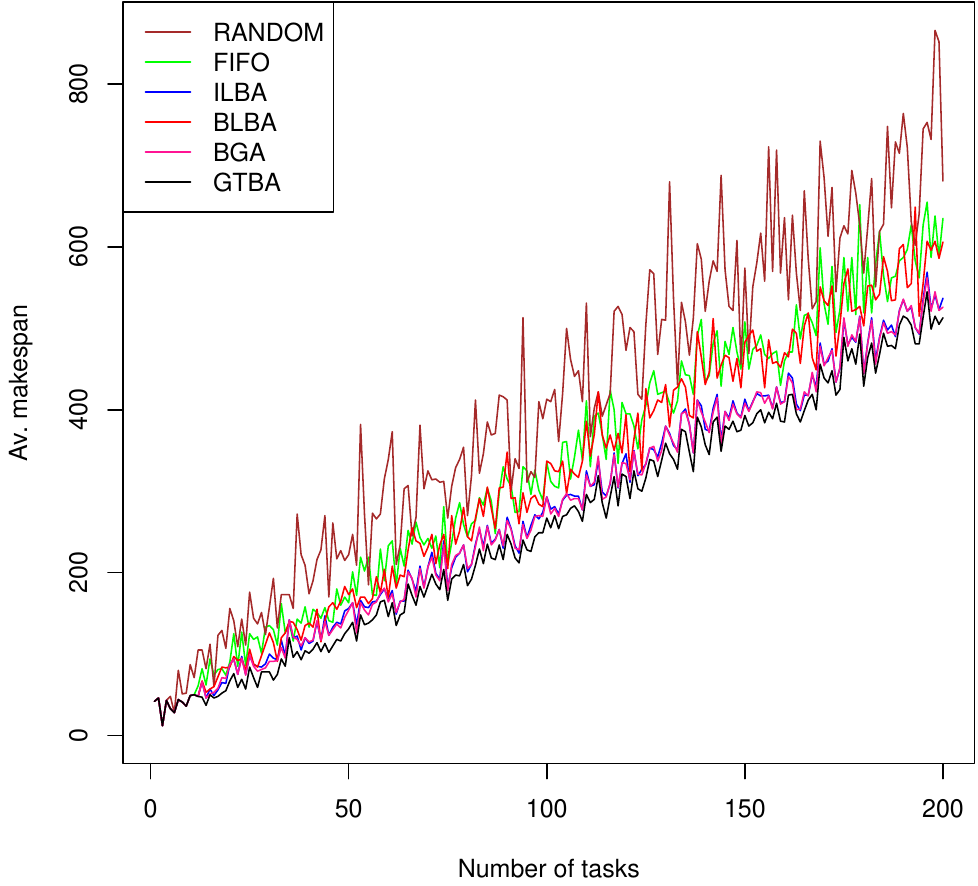}
\caption{Average makespan (in milliseconds) in a single time step.}
\label{fig1}
\end{figure}

As far as we know, most of the studies, such as \cite{Belal:2019,Kowsigan:2019} use random sampling, \cite{singh:2019,Kim:2019,Djigal:2019,Yu:2021} have a greedy nature, and studies such as \cite{Gulbaz:2021, Dasgupta:2013} use genetic algorithm with load balancer operator for minimizing the makespan and load balancing. So, for the comparison we are using:
\begin{itemize}
\item \textbf{Random algorithm: } The tasks are assigned randomly to virtual machines, \cite{Kowsigan:2019}.
\item \textbf{FIFO: } The tasks are assigned to the virtual machines based on their arrival order, \cite{Isard:2007}. 
\item \textbf{Genetic: } Tasks are assigned to virtual machines based on a genetic algorithm, then the balancer operator is used to balance the workload among virtual machines, called the balancer genetic algorithm (BGA), \cite{Gulbaz:2021}.
\item \textbf{Greedy: } The tasks are assigned to the virtual machines by the length of their queues. Two greedy methods are used: the method BLBA, in which each task is assigned to a virtual machine that has smallest number of scheduled tasks, and the method ILBA, in which each task is assigned to a virtual machine that requires the smallest time to complete all of its scheduled tasks, \cite{Yu:2021}.
\end{itemize}
We first assume that the task arrival rate (number of tasks received per second) is fixed and that the values vary between 1 and 200 tasks in a single time step. Given a fixed task arrival rate $n\in\{1,\ldots,200\}$, the new tasks are randomly selected 50 times according to their execution time (repetitions are allowed). For all 50 samples of tasks, we apply all task assignment algorithms and determine their average time with respect to the number of arrived tasks $n$. In Figure~\ref{fig1}, we show the average makespan of the different methods. In Figure~\ref{fig1}, the mean difference between the ILBA method and our method (GTBA) is $18.62(\pm7.44)$ and the mean difference between the BGA method and our method (GTBA) is $17.21(\pm7.17)$.

\begin{figure}[!t]\centering
\includegraphics[width=0.5\linewidth]{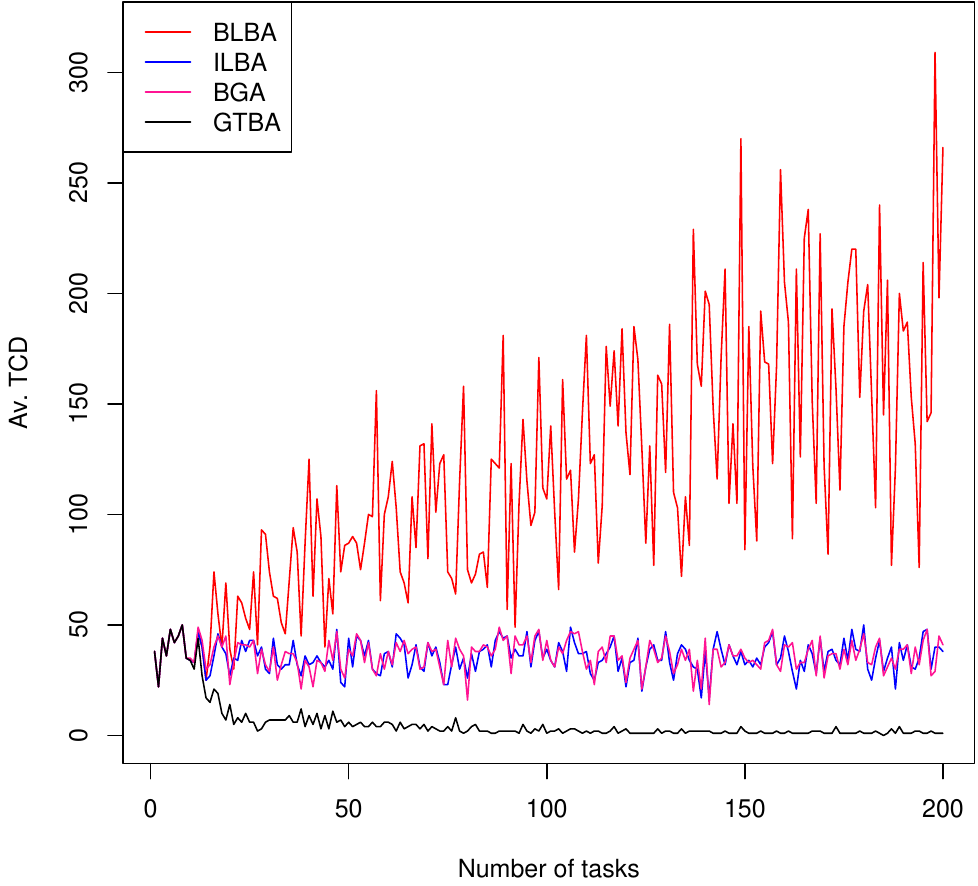}
\caption{$TCD$ (in milliseconds) as a function of number of arrival tasks in a single time step. It is possible to see that the ILBA, BLBA, and BGA have always larger $TCD$ compared with GTBA method.}
\label{fig3}
\end{figure}

One of the advantages of GTBA method is that all virtual machines complete their scheduled tasks at almost the same time. In other words, let $t_{\min}$ and $t_{\max}$ be the minimum time in which one of the virtual machines completes its scheduled tasks and the maximum total time in which all virtual machines complete all their scheduled tasks, respectively. Then we are interested in a task assignment to virtual machines such that the parameter, task completion difference, $TCD=t_{\max}-t_{\min}$, is small. In Figure~\ref{fig3} we show how $TCD$ changes as a function of the number of tasks arrived. Figures~\ref{fig1} and \ref{fig3} show that not only are the completion times of all tasks by virtual machines smaller, but also the completion times of all virtual machines are more balanced.

\begin{figure}[!t]\centering
\includegraphics[width=0.5\linewidth]{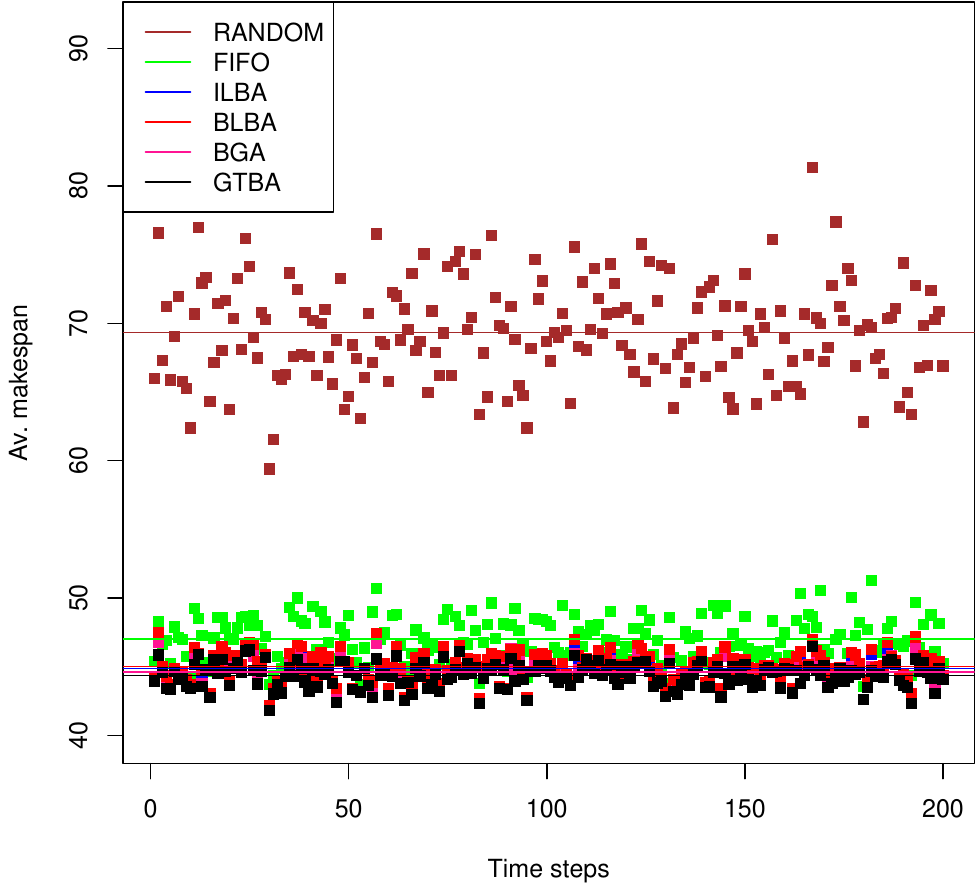}
\caption{Average makespan (in milliseconds) for each of the 200 time steps for $50$ experiments, where the number of arriving tasks follows a Poisson distribution with parameter $7$. The straight lines are the average makespans of all experiments for each method.}
\label{fig4}
\end{figure}

We now assume that the arrival rate of tasks changes at each time step (the total number of time steps is 200 and is a random variable from the Poisson distribution with parameter $7$, which means that in the long run on average $7$ tasks arrive in the system per second). In this setting we consider all virtual machines are free and do not have any scheduled task assigned to them when they receive the tasks assigned by methods. We repeat the same procedure $50$ times. First, we compute the average makespan, $TCD$, and the difference between GTBA method and the ILBA, BLBA, and BGA methods for each time step independently. The average makespan for each method is shown in Figure~\ref{fig4}, where we have averaged all the average makespans of all $50$ experiments at the end.
Now we take the average of the makespans of $50$ experiments, and run the same experiment $50$ more times. The mean of the average makespans of the different methods for all the $50$ runs are $69.65(\pm0.20)$, $47.19(\pm0.11)$, $44.91(\pm0.07)$, $45.06(\pm0.07)$, $44.66(\pm0.06)$, and $44.39(\pm0.05)$ for Random, Fifo, ILBA, BLBA, BGA, and GTBA, respectively. The mean difference between the ILBA method and our method (GTBA) is $0.52(\pm0.03)$ and the mean difference between the BGA method and our method (GTBA) is $0.27(\pm0.02)$. The mean $TCD$'s are $42.10(\pm0.06)$, $42.63(\pm0.07)$, $43.15(\pm0.07)$, and $40.93(\pm0.08)$ for ILBA, BLBA, BGA, and GTBA, respectively.
So, the tasks are better distributed among the virtual machines with GTBA method, and GTBA method reduces the average makespan. We found that ILBA performs better in load balancing compared to BLBA, which is also noted in \cite{Yu:2021}. Therefore, for a detailed comparison in the next experiment, we only consider ILBA.

\begin{figure}[!t]\centering
\includegraphics[width=0.5\linewidth]{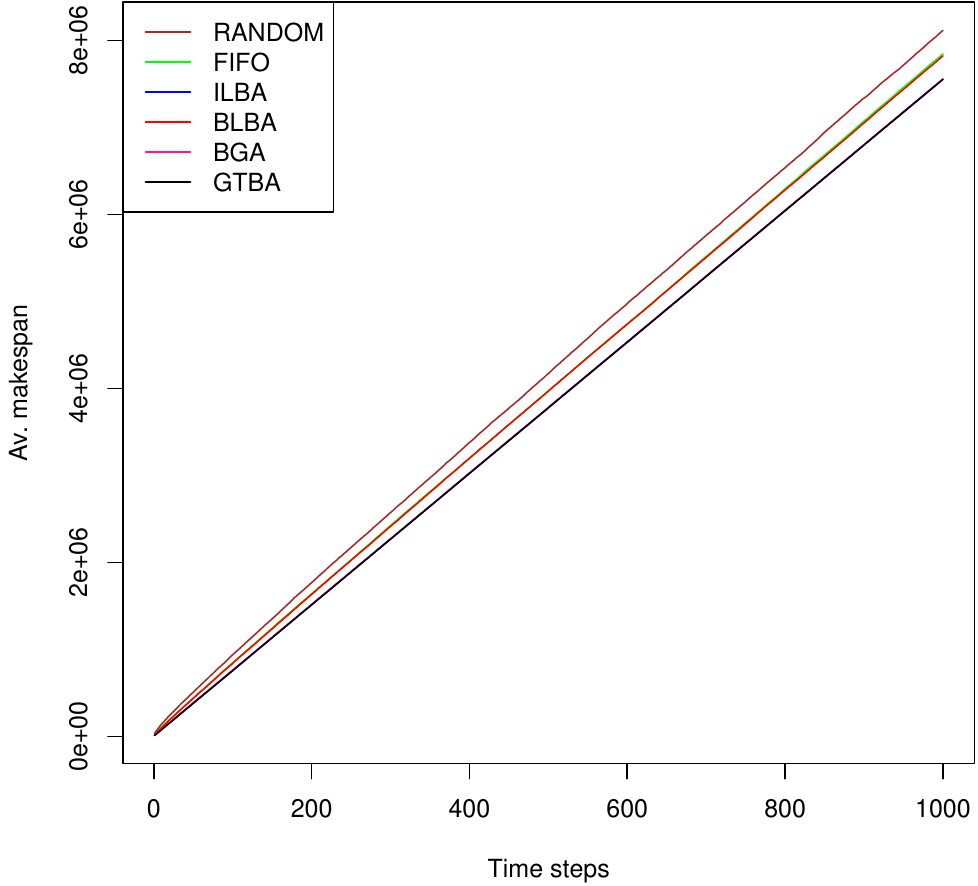}
\caption{Average makespan (in milliseconds) for the overall 200 time steps for $50$ experiments with $100$ virtual machines and 1000 time steps where the number of arriving tasks changes according to a Poisson distribution with parameter $150$.}
\label{fig8}
\end{figure}

\begin{figure}[!t]\centering
\includegraphics[width=0.5\linewidth]{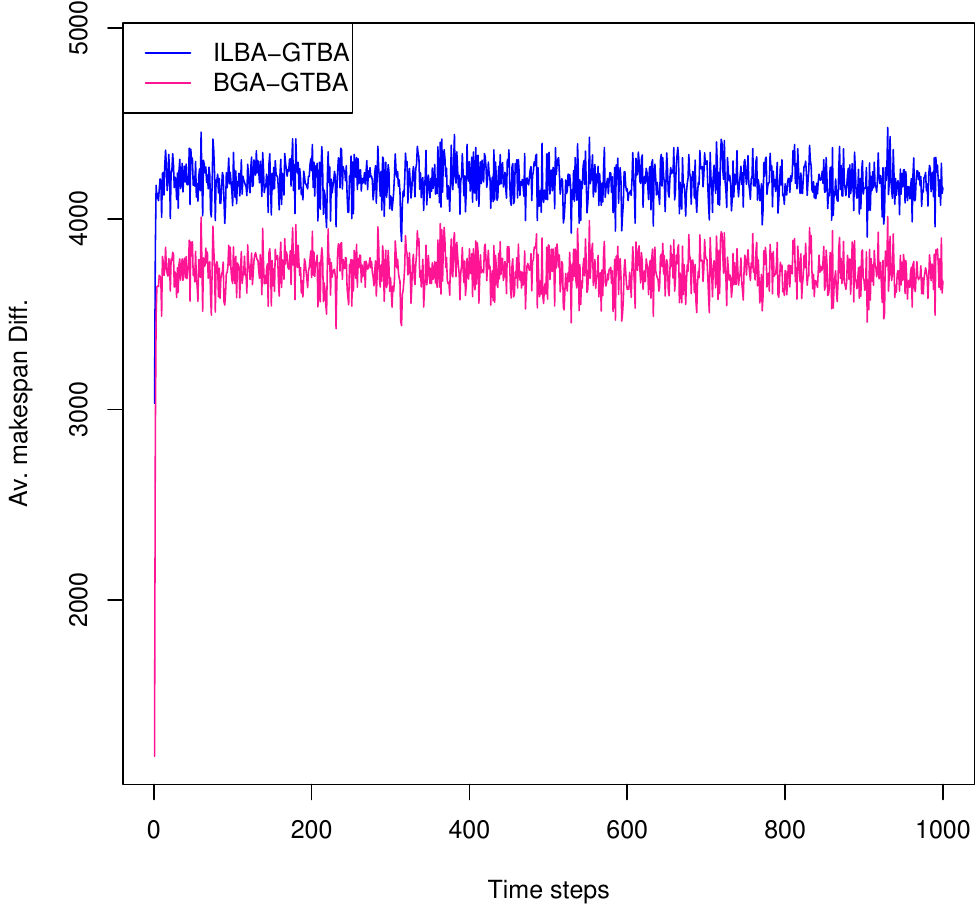}
\caption{Difference between the average makespans of the ILBA and BGA methods and GTBA method (in milliseconds) for the overall 1000 time steps. It can be seen that the ILBA and BGA methods always has a makespan larger or equal to GTBA method's.}
\label{fig9}
\end{figure}

As the final experiment, we consider $100$ virtual machines. We assume that the arrival rate of the tasks changes at each time step (the total time steps are 1000 and are a random variable from the Poisson distribution with parameter $150$). In this setting we measure the cummulative times of virtual machines for executing all their scheduled tasks at each time step within the total 1000 time steps. In this case, the virtual machimes might not be free when they receive the assigned tasks. We repeat the same procedure $50$ times. Now we calculate the average makespan, $TCD$ and the difference between GTBA method and the ILBA and BGA methods for all 1000 time steps, i.e. at each time step new tasks will be scheduled to virtual machines while virtual machines may already have some tasks to perform from previous time steps. The average makespan for each method is shown in Figure~\ref{fig8}, where we have averaged all the average makespans of all $50$ experiments at the end. The difference between the ILBA and BGA methods and GTBA method is not easily seen in this figure, so we present their differences in Figure~\ref{fig9}.
The average $TCD$ in each of 1000 time steps for all the methods does not have significant changes. The mean and standard deviation of the values for the average $TCD$ in each of 1000 time steps for the methods ILBA, BGA, and GTBA are $9111.59(\pm66.04)$, $9111.92(\pm65.83)$, and $2692.33(\pm113.79)$, respectively, showing that ILBA and BGA always has a larger average $TCD$ compared to GTBA method.

In the last experiment we measured the average time to completion of task scheduling after 1000 time steps using all methods. We first take the average time for 50 experiments and then add their values for 1000 time steps. The values are $241.5$, $68.4$, $1406.8$, $119.3$, $899.5$, and $236.6$ milliseconds for GTBA, Fifo, Random, BLBA, ILBA, and BGA, respectively. It shows that GTBA method is slightly slower than BGA and faster than ILBA but in both cases the average makespan is reduced by at least $3500$ milliseconds compared to ILBA and BGA, as shown in Figure~\ref{fig9}, which is a significant improvement compared to the execution time.

\begin{figure}[!t]\centering
\includegraphics[width=0.5\linewidth]{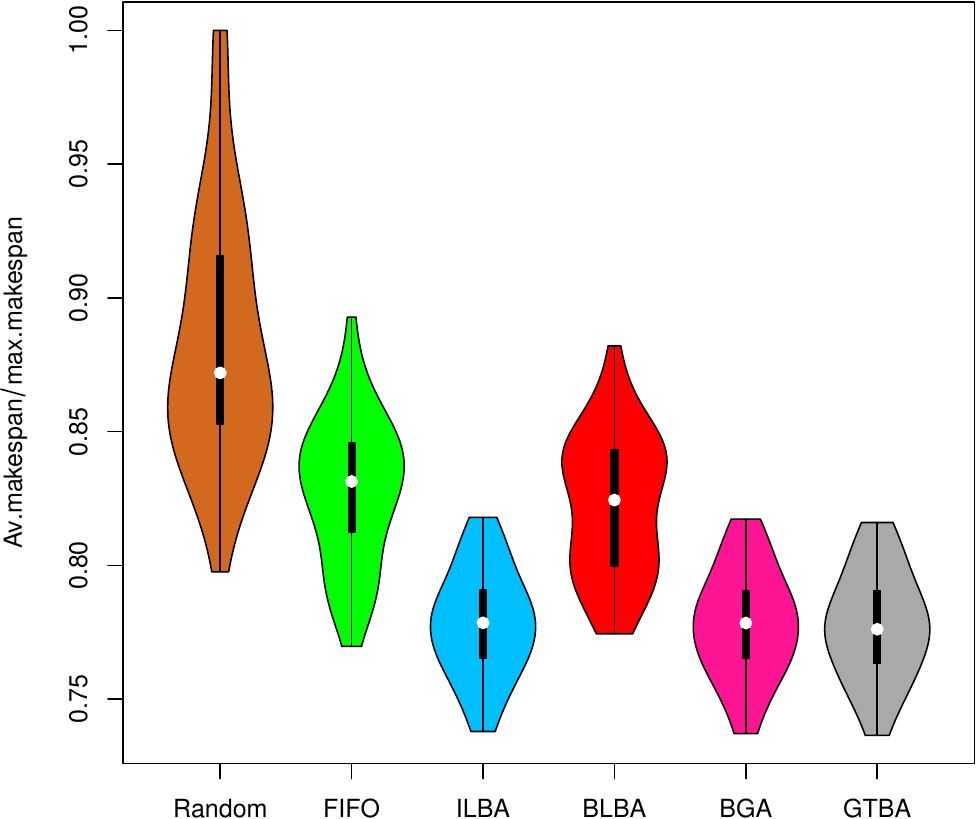}
\caption{Comparison of the kernel densities of different methods for time step 1000, where the average makespan is divided by the maximum average makespans of all methods.}
\label{figbar}
\end{figure}

\begin{figure}[!t]\centering
\includegraphics[width=0.5\linewidth]{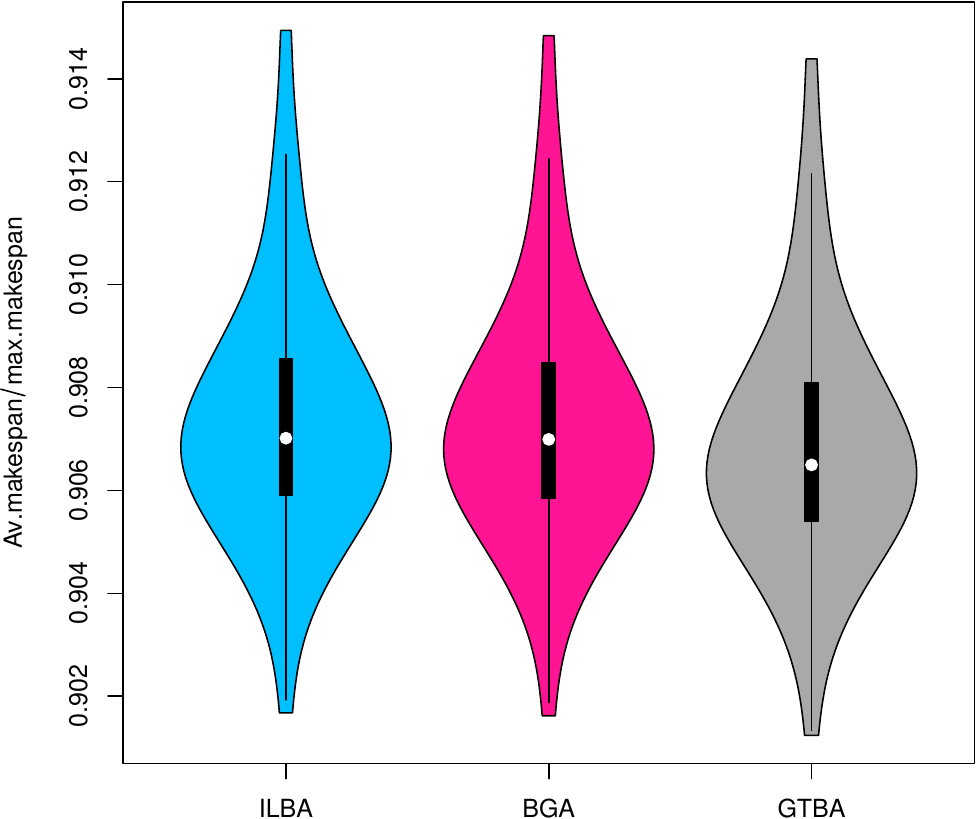}
\caption{Comparison of the kernel densities of the ILBA, BGA, and GTBA method for time step 1000, where the average makespan is divided by the maximum average makespans of all methods. It can be seen that GTBA method produces a smaller average makespan compared to ILBA and BGA.}
\label{figbarb}
\end{figure}

We compare the kernel densities of the different methods using the Violin plot for time step 1000. To better compare them, we divide the average makespan by the maximum average makespan of all methods, Figure~\ref{figbar}. Figure~\ref{figbarb} compares the kernel densities of methods ILBA, BGA, and GTBA for time step 1000. From Figure~\ref{figbarb}, it can be seen that GTBA method has a lower density for large average margins of the same height and a higher density for low average margins of the same height, being narrower at the top and wider at the bottom, i.e., GTBA method is more skewed to the top compared to the ILBA and BGA methods.

We also applied a $t$-test to the data obtained by the different methods for time step 1000 to test the difference between them. Table~\ref{tabttest} shows the value of $t$ in the $t$-test for the data obtained with the two methods along with the $p$-value. Looking at the first row of the Table~\ref{tabttest}, it is easy to observe that GTBA method is significantly different from all other methods because the absolute $t$-values are greater than $1.96$ and the $p$-values are less than $0.05$.
\begin{table*}[!t]
\caption{$t$-values and $p$-values in parentheses for the data obtained with Random, Fifo, ILBA, BLBA, BGA, and GTBA method compared to each other for time step 1000.}
\begin{center}
\resizebox{\linewidth}{!}{
\begin{tabular}{l|llllll}
&GTBA&Fifo&Random&ILBA&BLBA&BGA\\
\hline
GTBA&-&t = -45.49 (p $<$ 2.2e-16)&t = -40.23 (p $<$ 2.2e-16)&t = -36.10 (p $<$ 2.2e-16)&t = -47.19 (p $<$ 2.2e-16)&t = -30.24 (p = 1.70e-5)\\
Fifo&t = 45.49 (p $<$ 2.2e-16)&-&t = -16.88  (p $<$ 2.2e-16)&t = 44.70 (p $<$ 2.22e-16)&t = 2.77 (p = 0.0078)&t = 44.72 (p $<$ 2.2e-16)\\
Random&t = 40.23 (p $<$ 2.2e-16)&t = 16.88  (p $<$ 2.2e-16)&-&t = 39.93 (p $<$ 2.22e-16)&t = 19.31 (p $<$ 2.22e-16)&t = 39.97 (p $<$ 2.2e-16)\\
ILBA&t = 36.10 (p $<$ 2.2e-16)&t = -44.70 (p $<$ 2.22e-16)&t = -39.93 (p $<$ 2.22e-16)&-&t = -46.50 (p $<$ 2.2e-16)&t = 14.34 (p $<$ 2.2e-16)\\
BLBA&t = 47.19 (p $<$ 2.2e-16)&t = -2.77 (p = 0.0078)&t = -19.31 (p $<$ 2.22e-16)&t = 46.50 (p $<$ 2.2e-16)&-&t = 46.54 (p $<$ 2.22e-16)\\
BGA&t = 30.24 (p = 1.70e-5)&t = -44.72 (p $<$ 2.2e-16)&t = -39.97 (p $<$ 2.2e-16)&t = -14.34 (p $<$ 2.2e-16)&t = -46.54 (p $<$ 2.22e-16)&-
\end{tabular}}
\end{center}
\label{tabttest}
\end{table*}
\section{Complexity}
Let $m$ be the number of newly arrived tasks and $n$ the number of processors. The complexity of general dynamic scheduling can be easily determined as follows: Each task should be assigned to the appropriate processor. The total number of constraints on assigning tasks to different processing units is $mn$. Now, to find the optimal solution, we need to find the minimum makespan within these constraints that has complexity $O(mn)$. In GTBA method, the idea is to sort the tasks and processing units and then assign the tasks to the respective processor based on the obtained order. The complexity of GTBA method is then $O(\max\{m\log(m),n\log(n)\})$. The complexity of the genetic algorithm is $O(\max\{gnm\})$, where $g$ is the number of generations. And the complexity of the ILBA, which is only about finding the processing unit with minimum execution time, is $O(n)$.

\section{Scalability}
\begin{figure}[!t]\centering
\includegraphics[width=0.5\linewidth]{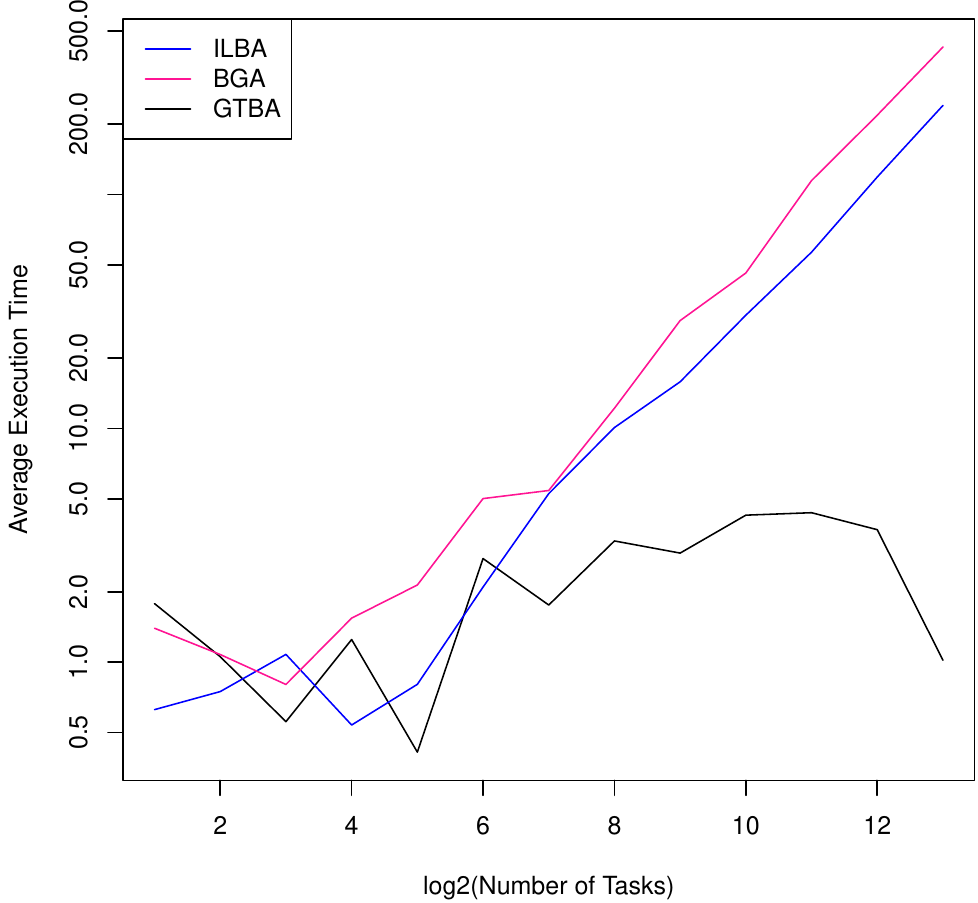}
\caption{The average execution time (in milliseconds) of methods ILBA (in blue), BGA (in red), and GTBA (in black) in setting 1 as functions of the average number of arrival tasks when the number of processing units is $256$ and the average number of newly arriving tasks is $2,4,\ldots,8192$. $x$ axis is in logarithmic, $\log_2$, scale.}
\label{scale1}
\end{figure}

\begin{figure}[!t]\centering
\includegraphics[width=0.5\linewidth]{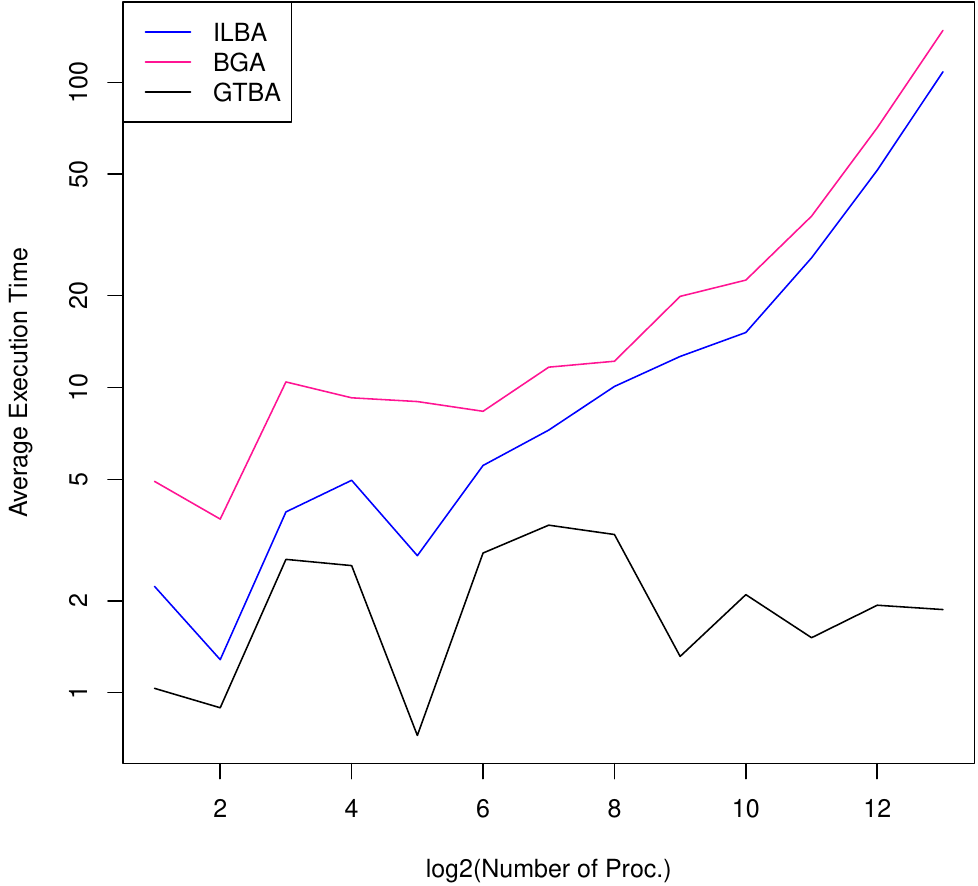}
\caption{The average execution time (in milliseconds) of methods ILBA (in blue), BGA (in red), and GTBA (in black) in setting 2 as functions of the number of processing units when the number of processing units is $2,4,\ldots,8192$ and the average number of arrival tasks is $256$. $x$ axis is in logarithmic, $\log_2$, scale.}
\label{scale1p}
\end{figure}

\begin{figure}[!t]\centering
\includegraphics[width=0.5\linewidth]{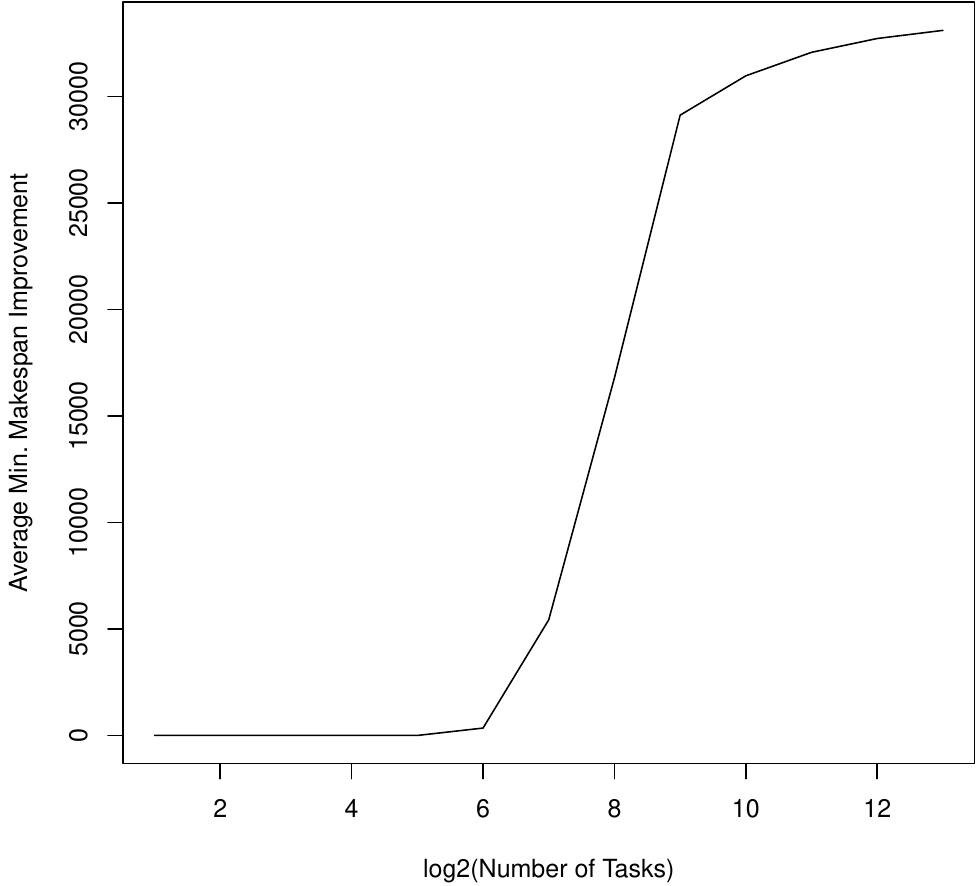}
\caption{The minimum improvement of the average makespan (in milliseconds) of BGA and ILBA compared with GTBA method in setting 1 as a function of the average number of arrival tasks. $x$ axis is in logarithmic, $\log_2$, scale.}
\label{scale2}
\end{figure}

\begin{figure}[!t]\centering
\includegraphics[width=0.5\linewidth]{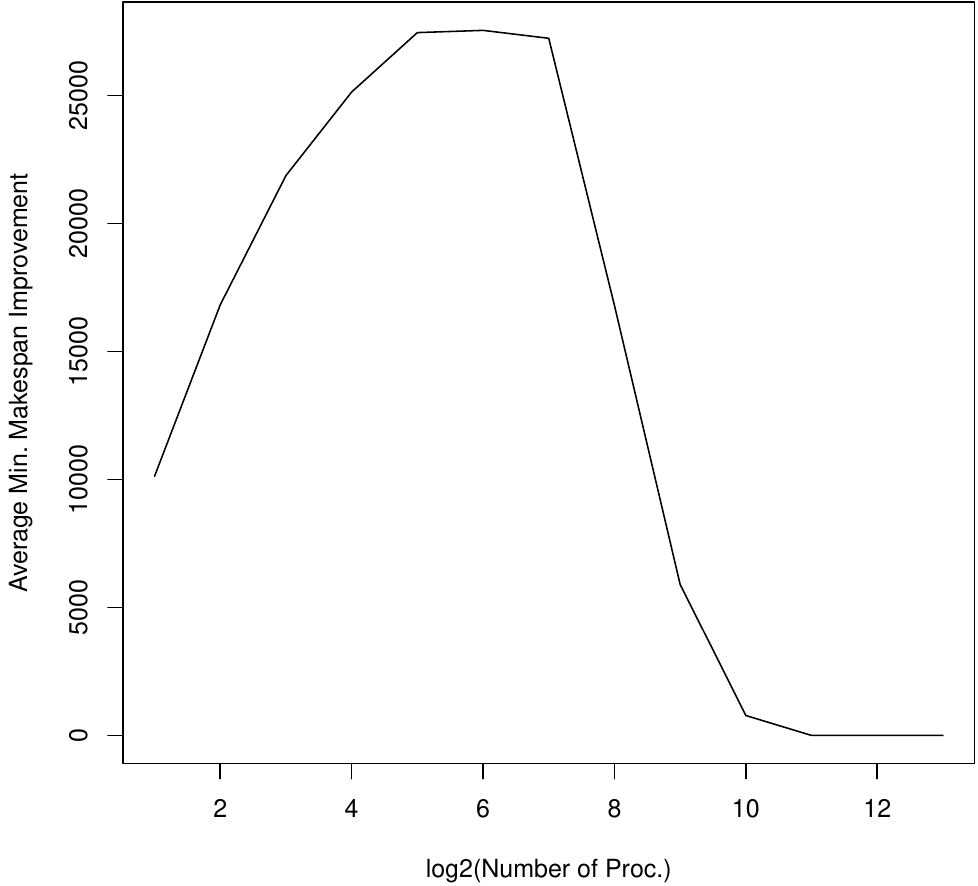}
\caption{The minimum improvement of the average makespan (in milliseconds) of BGA and ILBA compared with GTBA method in setting 2 as a function of the number of processing units. $x$ axis is in logarithmic, $\log_2$, scale.}
\label{scale2p}
\end{figure}

To better illustrate the scalability in two-dimensional space we consider two settings. We fix the number of processing units for \textbf{Setting 1} (fix average number of arrival tasks for \textbf{Setting 2}) equal to $256$, change the average number of arrival tasks (change the number of processing units) as $2,4,\ldots,8192$, and measure the average execution time (in milliseconds) of the ILBA, BGA, and GTBA methods as functions of the average number of arrival tasks (as functions of the number of processing units), setting 1, resluts in Figure \ref{scale1} (and setting 2, resluts in Figure \ref{scale1p}). For both settings we consider only the total time steps of $20$ and each experiment is repeated $50$ times. Figures \ref{scale2} and \ref{scale2p} show the minimum improvement of the average makespan of BGA and ILBA compared with GTBA method as a function of the average number of arrival tasks and the number of processing units for setting 1 and setting 2, respectively. Figures \ref{scale1} and \ref{scale2} show that GTBA has faster average execution time compared to ILBA and BGA. Figure \ref{scale1p} shows that for a fixed number of processing units, increasing the number of tasks further improves the average execution time of GTBA compared to BGA and ILBA. Figure \ref{scale2p} shows that for a fixed number of arrival tasks, increasing the number of processing units further improves the average makespan obtained by GTBA compared to BGA and ILBA for a small number of processing units. When the number of processing units is very large (much larger than the number of arrival tasks), the Pigeonhole principle suggests that each processing unit should be assigned at most one task to minimize the makespan so that all methods have the same makespan.
\section{Conclusion}
We proposed a method, GTBA, to assign the newly arrived tasks to different nodes such that the scheduled load of all nodes is balanced and obtains the minimum makespan.

We compared the GTBA method with existing scheduling methods. In GTBA, the differences in processing power and the total communication time are considered in the task time windows constraints. We assumed that tasks can only be performed by robots, and to perform each task, several algorithms need to be performed on the cloud infrastructure, on the fog, or on the robots. We can restrict the proposed method such that the algorithms are performed only on the cloud or on the fog. Then, the solution will minimize the makespan with load balancing on the virtual machines of the cloud infrastructure. In addition, we may assume that some of the tasks can only be performed by some of the nodes. In this case, we can restrict streams, in the sense that the sequence of all tasks in each stream can be performed by at least one node. In other words, the method can distinguish the tasks that can be performed by certain nodes and then construct the grid of all tasks by first identifying the relative streams for these tasks and then extend it to all the tasks. Then, task scheduling can be done by assigning each stream to a node that is capable to perform it, and the streams are assigned to nodes according to Proposition~\ref{prop:prop1} and Corollary~\ref{cor:cor1}. GTBA has a greedy nature, but the main difference against other greedy methods is that, as we have proved, GTBA is the optimal way of load balancing while minimizing the makespan.

The proof we provide serves as a guarantee of performance, and the analysis of the algorithm is part of the proof. The experimental section is used to illustrate how the proposed method works in randomly generated settings. We show that GTBA outperforms greedy algorithms for load balancing, which means that the result of \cite{Buchem:2021} can be improved if we replace the greedy algorithm with GTBA as policies for load balancing. We proved the correctness of GTBA method and illustrated its performance compared to five other methods through simulations, which showed that it outperforms all of them. For future work, we will include communication instability \cite{Ours:2020h} and possible node failure, for more realistic scenarios.


\section*{Acknowledgment}

This work was supported by operation Centro-01-0145-FEDER-000019 - C4 - Centro de Compet\^{e}ncias em Cloud Computing, cofinanced by the European Regional Development Fund (ERDF) through the Programa Operacional Regional do Centro (Centro 2020), in the scope of the Sistema de Apoio \`{a} Investiga\c{c}\~{a}o Cientif\'{i}ca e Tecnol\'{o}gica - Programas Integrados de IC\&DT. This work was supported by NOVA LINCS (UIDB/04516/2020) with the financial support of FCT-Funda\c{c}\~{a}o para a Ci\^{e}ncia e a Tecnologia, through national funds.

\bibliographystyle{unsrt}
\bibliography{sample}

\end{document}